\documentclass{article}
\usepackage{ijcai11}
\usepackage{times}

\usepackage[noend]{algpseudocode}
\usepackage[ruled]{algorithm}
\usepackage{url}
\usepackage{framed}
\usepackage{amsfonts,amsmath,amsthm}
\usepackage{graphicx}
\usepackage{url}
\usepackage{color}

\newcommand {\mean} {\ensuremath {\mathop{\mathrm{mean}}}}
\newcommand {\median} {\ensuremath {\mathop{\mathrm{median}}}}

\newcommand {\IE} {\ensuremath {\mathbb{E}}}

\title{Rational Deployment of CSP Heuristics}
\author {David Tolpin \and Solomon Eyal Shimony \\
Ben-Gurion University of the Negev, Beer Sheva, Israel \\
\{tolpin,shimony\}@cs.bgu.ac.il}

\begin{document}

\maketitle

\begin{abstract}

Heuristics are crucial tools in decreasing search effort in varied
fields of AI. In order to be effective, a heuristic must be efficient
to compute, as well as provide useful information to the search
algorithm.  However, some well-known heuristics which do well in
reducing backtracking are so heavy that the gain of
deploying them in a search algorithm might be outweighed by their overhead.
 
We propose a rational metareasoning approach to decide when
to deploy heuristics, using CSP backtracking search as a case study.  In
particular, a value of information approach is taken to adaptive
deployment of solution-count estimation heuristics for value ordering.
Empirical results show that indeed the proposed mechanism successfully
balances the tradeoff between decreasing backtracking and heuristic
computational overhead, resulting in a significant overall search time
reduction.

\end{abstract}

\section{Introduction}

Large search spaces are common in artificial intelligence, heuristics being
of major importance in limiting search efforts.
The role of a heuristic, depending on type of search algorithm,
is to decrease the number of nodes expanded (e.g. in A* search),
the number of candidate actions considered (planning), or
the number of backtracks in constraint satisfaction problem (CSP) solvers.
Nevertheless, some sophisticated heuristics have considerable
computational overhead, significantly decreasing their overall effect
\cite{HorschHavens.pac,Kask.solcount},
even causing {\em increased} total runtime in pathological cases.
It has been recognized that control of this overhead can be essential
to improve search performance; e.g. by selecting which heuristics to evaluate
in a manner dependent on the state of the search \cite{Wallace.macheur,Karpas.maxornot}.

We propose a rational metareasoning approach \cite{Russell.right} to decide
when and how to deploy heuristics, using CSP backtracking search as a case study.
The heuristics examined are various {\em solution count estimate} heuristics for
value ordering \cite{Meisels.solcount,HorschHavens.pac}, which
are expensive to compute, but can significantly decrease the number of
backtracks. These heuristics make a good case study, as their overall utility, taking computational
overhead into account, is sometimes detrimental; and yet,
by employing these heuristics adaptively, it may still be possible to achieve an
overall runtime improvement, even in these pathological cases.
Following the metareasoning approach, the value of information (VOI) of
a heuristic is defined in terms of total search time saved, and the
heuristic is computed such that the expected net VOI is positive.

We begin with background on metareasoning and CSP (Section \ref{sec:background}), 
followed by a re-statement of value ordering in terms
of rational metareasoning (Section \ref{sec:generic}), allowing a definition of VOI
of a value-ordering heuristics --- a contribution of this paper.
This scheme is then instantiated to handle
our case-study of backtracking search in CSP (Section \ref{sec:csp-rational}),
with parameters specific to value-ordering
heuristics based on solution-count estimates, the main contribution of this paper.
Empirical results (Section \ref{sec:empirical})
show that the proposed mechanism successfully balances the tradeoff
between decreasing backtracking and
heuristic computational overhead, resulting in a significant overall search time reduction.
Other aspects of such tradeoffs are also analyzed empirically.
Finally, possible future extensions of the proposed mechanism are discussed
(Section \ref{sec:discussion}), as well as an examination of related work.

\section{Background}
\label{sec:background}

\subsection{Rational metareasoning}
\label{sec:ratimeta}

In rational metareasoning \cite{Russell.right},
a problem-solving agent can perform {\em base-level} actions from a known
set $\{A_i\}$. Before committing to an action, the agent may perform a sequence of
{\em meta-level} ``deliberation'' actions from a set $\{S_j\}$. At any given
time there is an ``optimal'' base-level action, $A_\alpha$, that maximizes the agent's expected
utility:
\begin{equation}
\label{eq:ratimeta-eu}
\alpha=\arg \max_i \sum_k P(W_k) U(A_i,W_k)
\end{equation}
where $\{W_k\}$ is the set of possible world states, $U(A_i, W_k)$ is
the utility of performing action $A_i$ in state $W_k$, and $P(W_k)$ is
the probability that the current world state is $W_k$.

A meta-level action provides information and affects the choice of
the base-level action $A_\alpha$. The {\em value of information} (VOI) of a meta-level
action $S_j$ is the expected difference between the expected utility
of $S_j$ and the expected utility of the current $A_\alpha$, where $P$
is the current belief distribution about the state of world, and $P^j$
is the belief-state distribution of the agent after the
computational action $S_j$ is performed, given the outcome of $S_j$:
\begin{equation}
\label{eq:ratimeta-netvoi}
V(S_j)=\IE_P(\IE_{P^j}(U(S_j))-\IE_{P^j}(U(A_\alpha)))
\end{equation}
Under certain assumptions, it is possible to capture the dependence of
utility on time in a separate notion of {\em time cost} $C$. Then, 
Equation (\ref{eq:ratimeta-netvoi}) can be rewritten as:
\begin{equation}
\label{eq:v=bc}
V(S_j)=\Lambda(S_j)-C(S_j)
\end{equation}
where the {\em intrinsic value of information}
\begin{equation}
\label{eq:bg-limited-benefit}
\Lambda(S_j)=\IE_P\left(\IE_{P^j}(U(A_\alpha^j))-\IE_{P^j}(U(A_\alpha))\right)
\end{equation}
is the expected difference between the {\em intrinsic expected utilities} of the new
and the old selected base-level action, computed after the meta-level
action is taken.

\subsection{Constraint satisfaction}

A constraint satisfaction problem (CSP) is defined by a set
of variables $\mathcal{X}=\{X_1, X_2, ...\}$, and a set of
constraints $\mathcal{C}=\{C_1, C_2, ...\}$. Each variable $X_i$ has a non-empty domain
$D_i$ of possible values. Each constraint $C_i$ involves some subset
of the variables---the scope of the constraint--- and specifies the
allowable combinations of values for that subset. An assignment that
does not violate any constraints is called {\em consistent} (or {\em a solution}).
There are numerous variants of CSP settings and algorithmic paradigms. This
paper focuses on binary CSPs over discrete-values variables,
and backtracking search algorithms \cite{Tsang.csp}.

A basic method used in numerous CSP search algorithm is
that of maintaining arc consistency (MAC)
\cite{Sabin.mac}. There are several versions of MAC; all
share the common notion of {\em
arc consistency}.
A variable $X_i$ is arc-consistent with $X_j$ if for every value $a$ of
$X_i$ from the domain $D_i$ there is a value $b$ of $X_j$ from the
domain $D_j$ satisfying the constraint between $X_i$ and $X_j$. MAC
maintains arc consistency for all pairs of variables, and
speeds up backtracking search by pruning many inconsistent
branches.

CSP backtracking search algorithms typically employ
both variable ordering \cite{Tsang.csp} and value ordering heuristics.
The latter type include \emph{minimum conflicts} \cite{Tsang.csp}, which
orders values by the number of conflicts they cause
with unassigned variables, \emph{Geelen's promise}
\cite{Geelen.promise} --- by the product of domain sizes,
and \emph{minimum impact}
\cite{Refalo.impact} orders values by relative impact of the value
assignment on the product of the domain sizes.

Some value-ordering heuristics are based on solution count estimates
\cite{Meisels.solcount,HorschHavens.pac,Kask.solcount}: solution
counts for each value assignment of the current variable are
estimated, and assignments (branches) with the greatest solution count
are searched first.  The heuristics are based on the assumption that
the estimates are correlated with the true number of solutions, and
thus a greater solution count estimate means a higher probability that
a solution be found in a branch, as well as a shorter search time to
find the first solution if one exists in that
branch. \cite{Meisels.solcount} estimate solution counts by
approximating marginal probabilities in a Bayesian network derived
from the constraint graph; \cite{HorschHavens.pac} propose the
\emph{probabilistic arc consistency} heuristic (pAC) based on
iterative belief propagation for a better accuracy of relative
solution count estimates; \cite{Kask.solcount} adapt Iterative
Join-Graph Propagation to solution counting, allowing a tradeoff
between accuracy and complexity. These methods vary by computation
time and precision, although all are rather computationally
heavy. Principles of rational metareasoning can be applied
independently of the choice of implementation, to decide when
to deploy these heuristics.

\section{Rational Value-Ordering}
\label{sec:generic}

The role of (dynamic) value-ordering is to determine the order of
values to assign to a variable $X_k$ from its domain $D_k$, at a
search state where values have already been assigned to $(X_1, ...,
X_{k-1})$. We make the standard assumption that the ordering may
depend on the search state, but is not re-computed as a result of
backtracking from the initial value assignments to $X_k$: a new
ordering is considered only after backtracking up the search tree
above $X_k$.

Value ordering heuristics provide information on future search
efforts, which can be summarized by 2 parameters:
\begin{itemize}
\item  $T_i$---the expected time to find a solution containing
  assignment  $X_k=y_{ki}$ or verify that there are no such solutions;
\item  $p_i$---the ``backtracking probability'', that there will be no solution
consistent with $X_k=y_{ki}$.
\end{itemize}
These are treated as the algorithm's subjective probabilities about future search
in the current problem instance, rather than actual distributions over problem instances.
Assuming correct values of these parameters, and independence of backtracks,
the expected remaining search time in the subtree under $X_k$ for ordering $\omega$ is given by:
\begin{equation}
\label{eq:expected-search-time}
T^{s|\omega}=T_{\omega(1)}+\sum_{i=2}^{|D_k|}T_{\omega(i)}\prod_{j=1}^{i-1}p_{\omega(j)}
\end{equation}
In terms of rational metareasoning, the ``current'' optimal base-level action is picking
the $\omega $ which optimizes $T^{s|\omega}$.
Based on a general property of functions on sequences \cite{MonmaSidney.sequencing}, it can
be shown that $T^{s|\omega}$ is minimal if 
the values are sorted  by increasing order of $\frac {T_i} {1-p_i}$.

A candidate heuristic $H$ (with computation time $T^H$)  generates
an ordering by providing an updated (hopefully more precise)
value of the parameters $T_i, p_i$ for value assignments
$X_k=y_{ki}$, which may lead to a new optimal ordering $\omega_H$,
corresponding to a new base-level action.  The total expected
remaining search time is given by:

\begin{equation}
\label{eq:net-expected-time}
T=T^H+E[T^{s|\omega_H}]
\end{equation}

Since both $T^H$ (the ``time cost'' of $H$ in metareasoning terms)
and $T^{s|\omega_H}$ contribute to $T$, even a heuristic that
improves the estimates and ordering may not be useful.  It may be better not
to deploy $H$ at all, or to update $T_i, p_i$ only for some of the assignments.
According to the rational metareasoning approach (Section~\ref{sec:ratimeta}),
the intrinsic VOI $\Lambda_i$ of estimating $T_i, p_i$ for the $i$th assignment
is the expected decrease in the expected search time:
\begin{equation}
\label{eq:value-ordering-benefit}
\Lambda_i=\IE\left[T^{s|\omega_-}-T^{s|\omega_{+i}}\right]
\end{equation}
where $\omega_-$ is the optimal ordering based on priors,
and $\omega_{+i}$ on values after updating $T_i, p_i$.
Computing new estimates (with overhead $T^c$) for values $T_i, p_i$
is beneficial just when the net VOI is positive:
\begin{equation}
\label{eq:voi}
V_i=\Lambda_i-T^c
\end{equation}
To simplify estimation of $\Lambda_i$, the expected
search time of an ordering is estimated as though the parameters are
computed only for $\omega_-(1)$ (essentially the metareasoning 
subtree independence assumption). Other value assignments are assumed
to have the prior (``default'') parameters $T_\mathrm{def}, p_\mathrm{def}$.
Assume w.l.o.g. that $\omega_-(1)=1$:
\begin{equation}
\label{eq:time-estimate}
T^{s|\omega_-}=T_1+p_1\sum_{i=2}^{|D_k|}T_\mathrm{def}p_\mathrm{def}^{i-2}
   =T_1+p_1T_\mathrm{def}\frac{1-p_\mathrm{def}^{(|D_k|-1)}} {1-p_\mathrm{def}}
\end{equation}
and the intrinsic VOI of the $i$th deliberation action is:
\begin{equation}
\label{eq:benefit-estimate}
\Lambda_i=\IE\left[G(T_i, p_i)\,\Big|\;\frac {T_i} {1-p_i} < \frac {T_1} {1-p_1}\right]
\end{equation}
where $G(T_i, p_i)$ is the search time gain given the heuristically computed values $T_i, p_i$:
\begin{equation}
\label{eq:gain}
G(T_i, p_i) = T_1-T_i+(p_1-p_i)T_\mathrm{def}\frac{1-p_\mathrm{def}^{(|D_k|-1)}}{1-p_\mathrm{def}}
\end{equation}
In some cases, $H$ provides estimates only for the expected
search time $T_i$. In such cases, the backtracking probability $p_i$
can be bounded by the Markov inequality as the probability for the
given assignment that the time $t$ to find a solution or verify that
no solution exists is at least the time $T_i^{all}$ to find all
solutions: $p_i=P\left(t\ge T_i^{all}\right)\le \frac{T_i} {T_i^{all}}$, 
and the bound can be used as the probability estimate:
\begin{equation}
\label{eq:backtracking-probability-estimate}
p_i\approx\frac{T_i} {T_i^{all}}
\end{equation}

Furthermore, note that in harder problems  the probability of
backtracking from variable $X_k$ is proportional to $p_\mathrm{def}^{(|D_k|-1)}$, and it is
reasonable to assume that backtracking probabilities above $X_k$
(trying values for $X_1, ..., X_{k-1}$)  are still significantly greater than 0.
Thus, the ``default'' backtracking
probability $p_\mathrm{def}$ is close to 1, and consequently:
\begin{equation}
  \label{eq:p-approx-one-estimate}
T_i^{all} \approx T_\mathrm{def},\quad\frac{1-p_\mathrm{def}^{(|D_k|-1)}}{1-p_\mathrm{def}} \approx |D_k|-1
\end{equation}
By substituting (\ref{eq:backtracking-probability-estimate}),
(\ref{eq:p-approx-one-estimate}) into (\ref{eq:gain}),
estimate (\ref{eq:gain-estimate}) for $G(T_i, p_i)$ is obtained:
\begin{eqnarray}
\label{eq:gain-estimate}
G(T_i, p_i)&\approx&T_1-T_i+(\frac {T_1} {T_1^{all}}-\frac {T_i} {T_i^{all}})T_\mathrm{def}\frac{1-p_\mathrm{def}^{(|D_k|-1)}}{1-p_\mathrm{def}}\nonumber\\
           &\approx&(T_1-T_i)|D_k|
\end{eqnarray}
Finally, since (\ref{eq:backtracking-probability-estimate}), (\ref{eq:p-approx-one-estimate}) imply that $T_i<T_1 \Leftrightarrow \frac {T_i} {1-p_i} < \frac {T_1} {1-p_1}$, 
\begin{eqnarray}
\label{eq:benefit-superestimate}
\Lambda_i\approx\IE\left[(T_1-T_i)|D_k|\,\Big|\; T_i<T_1 \right]
\end{eqnarray}

\section{VOI of Solution Count Estimates}
\label{sec:csp-rational}

The estimated solution count for an assignment may be used to estimate
the expected time to find a solution for the assignment under the
following assumptions\footnote{We do not claim
that this is a valid model of CSP search; rather, we argue that even with such a crude model
one can get significant runtime improvements.}:
\begin{enumerate}
\item Solutions are roughly evenly distributed in the search space, that is,
   the distribution of time to find a solution can be modeled by a
   Poisson process.
\item Finding all solutions for an assignment $X_{k}=y_{ki}$
takes roughly the same time for all assignments to the variable $X_k$. Prior work
   \cite{Meisels.solcount,Kask.solcount} demonstrates that
   ignoring the differences in subproblem sizes is justified.
\item The expected time to find all solutions for an assignment
  divided by its solution count estimate is a
  reasonable estimate for the expected time to find a single solution. 
\end{enumerate}
Based on these assumptions, $T_i$ can be estimated as $\frac {T^{all}}
{|D_k|n_i}$ where $T^{all}$ is the expected time to find all
solutions for all values of $X_k$, and $n_i$ is the
solution count estimate for $y_{ki}$; likewise, $T_1=\frac
{T^{all}} {|D_k|n_\mathrm{max}}$, where $n_\mathrm{max}$ is the currently
greatest $n_i$.  By substituting the expressions for
$T_i$, $T_1$ into (\ref{eq:benefit-superestimate}), obtain
as the intrinsic VOI of computing $n_i$:
\begin{equation}
\label{eq:benefit-estimate-sc}
\Lambda_i=T^{all} \sum_{n=n_\mathrm{max}}^\infty\left(
  \frac 1 {n_\mathrm{max}} - \frac 1 n\right) P(n, \nu)
\end{equation}
where $P(n, \nu)=e^{-\nu}\frac {\nu^n} {n!}$ is the probability,
according to the Poisson distribution, to find $n$ solutions for a
particular assignment when the mean number of solutions per assignment
is $\nu=\frac N {|D_k|}$, and $N$ is the estimated solution count for
all values of $X_k$, computed at an earlier stage of the algorithm.

Neither $T^{all}$ nor $T^c$, the time to estimate the solution count
for an assignment, are known. However, for relatively low solution
counts, when an invocation of the heuristic has high intrinsic VOI, both
$T^{all}$ and $T^c$ are mostly determined by the time spent eliminating
non-solutions. Therefore, $T^c$ can be assumed approximately proportional
to $\frac {T^{all}} {|D_k|}$, the average time to find all solutions
for a single assignment, with an unknown factor $\gamma < 1$.

\begin{equation}
\label{eq:csp-gamma}
T^c \approx \gamma \frac {T^{all}} {|D_k|}
\end{equation}
Then, $T^{all}$ can be eliminated from both $T^c$ and $\Lambda$. 
Following Equation~(\ref{eq:voi}), the solution count 
should be estimated whenever the net VOI is positive:
\begin{equation}
\label{eq:csp-solcount-condition}
V(n_\mathrm{max}) \propto |D_k|e^{-\nu}\sum_{n=n_\mathrm{max}}^\infty \! \! \left( \frac 1 {n_\mathrm{max}} - \frac 1 n\right) \frac {\nu^n} {n!}-\gamma
\end{equation}
The infinite series in
(\ref{eq:csp-solcount-condition}) rapidly converges, and an
approximation of the sum can be computed efficiently. As done
in Section \ref{sec:empirical}, $\gamma$ can be learned
offline from a set of problem instances of a certain kind for the
given implementation of the search algorithm and the solution counting
heuristic.

Algorithm~\ref{alg:csp-solcount} implements rational value ordering.
The procedure receives problem
instance $csp$ with assigned values for variables $X_1, ..., X_{k-1}$,
variable $X_k$ to be ordered, and estimate $N$ of the
number of solutions of the problem instance
(line~\ref{alg:csp-solcount-args}); $N$ is computed at the previous
step of the backtracking algorithm as the solution count estimate for the chosen
assignment for $X_{k-1}$, or, if $k=1$, at the beginning of the search
as the total solution count estimate for the instance. Solution counts
estimates $n_i$ for some of the assignments are re-computed
(lines~\ref{alg:csp-solcount-while}--\ref{alg:csp-solcount-endwhile}),
and then the domain of $X_k$, ordered by non-increasing solution count
estimates of value assignments, is returned
(lines~\ref{alg:csp-solcount-sort}--\ref{alg:csp-solcount-return}).

\begin{algorithm}[h]
\caption{Value Ordering via Solution Count Estimation}
\label{alg:csp-solcount}
\begin{algorithmic}[1]
\Procedure{ValueOrdering-SC}{$csp, X_k, N$}\label{alg:csp-solcount-args}
  \State $D \gets D_k$,\hspace{1em}$n_\mathrm{max} \gets \frac N {|D|}$
  \State {\bf for all} {$i$ {\bf in} $1..|D|$} {\bf do} $n_i \gets n_\mathrm{max}$
  \While {$V(n_\mathrm{max})>0$} \label{alg:csp-solcount-while}
    \State choose $y_{ki} \in D$ arbitrarily
    \State $D \gets D \setminus \{y_{ki}\}$
    \State $csp' \gets csp$ with $D_k=\{y_{ki}\}$
    \State $n_i \gets$ {\sc EstimateSolutionCount}($csp'$)
    \State {\bf if} {$n_i>n_\mathrm{max}$} {\bf then} $n_\mathrm{max} \gets n_i$
  \EndWhile \label{alg:csp-solcount-endwhile}
  \State {\bf end while}
  \State $D_{ord} \gets$ sort $D_k$ by non-increasing $n_i$ \label{alg:csp-solcount-sort}
  \State {\bf return} $D_{ord}$ \label{alg:csp-solcount-return}
\EndProcedure
\end{algorithmic}
\end{algorithm}

\section{Empirical Evaluation}
\label{sec:empirical}


Specifying the algorithm parameter $\gamma$ is the first issue.
$\gamma$ should be a characteristic of the
implementation of the search algorithm, rather than of the problem
instance; it is also desirable that the performance of the algorithm not
be too sensitive to fine tuning of this parameter.


Most of the  experiments were conducted on sets of random problem instances
generated according to Model RB \cite{Xu.rb}. The empirical evaluation
was performed in two stages. In the first stage, several benchmarks
were solved for a wide range of values of $\gamma$, and an appropriate
value for $\gamma$ was chosen. In the second stage, the search was run
on two sets of problem instances with the chosen $\gamma$, as well as
with exhaustive deployment, and with the minimum conflicts
heuristic, and the search time distributions were compared for each of
the value ordering heuristics.

The AC-3 version of MAC was used for the experiments,
with some modifications \cite{Sabin.mac}. Variables were ordered using
the maximum degree variable ordering heuristic.\footnote{A dynamic
variable ordering heuristic, such as dom/deg, may result in shorter search times in
general, but gave no significant improvement in our experiments; on
the other hand, static variable ordering simplifies the analysis.} The solution counting heuristic
was based on the solution count estimate proposed in
\cite{Meisels.solcount}. The source code is available from \url{http://ftp.davidashen.net/vsc.tar.gz}.

\subsection{Benchmarks}

\begin{figure}[h]
\centering
\includegraphics[scale=0.67]{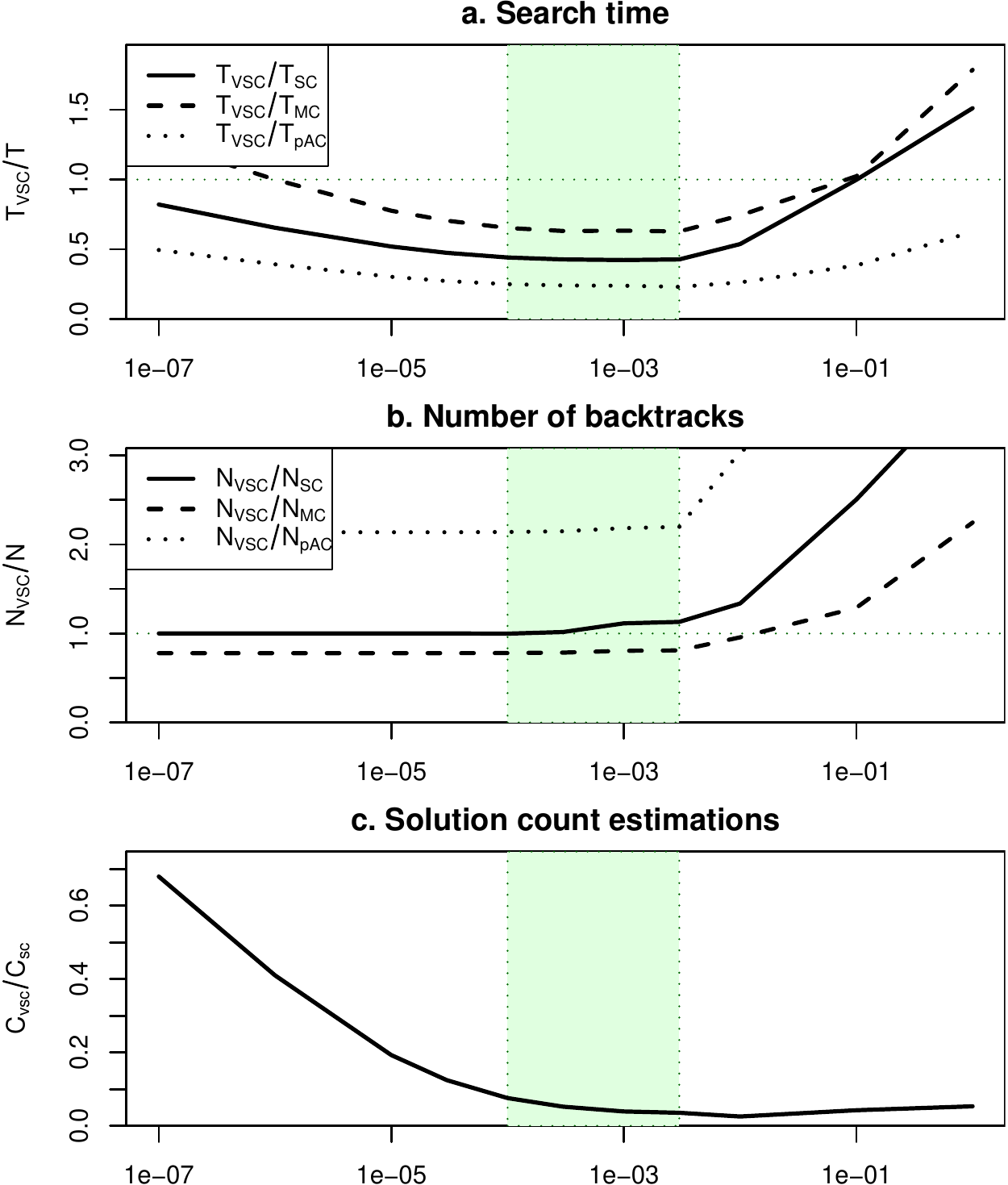}
\caption{Influence of $\gamma$ in CSP benchmarks}
\label{fig:benchmarks}
\end{figure}

CSP benchmarks from CSP Solver Competition~2005
\cite{Boussemart.benchmarks} were used. 14 out of 26 benchmarks
solved by at least one of the solvers submitted for the competition
could be solved with 30 minutes timeout by the solver used for this
empirical study for all values of $\gamma$: $\gamma=0$ and the
exponential range $\gamma \in \{10^{-7}, 10^{-6}, ..., 1\}$, as well
as with the minimum-conflicts heuristic and the pAC heuristic.

Figure~\ref{fig:benchmarks}.a shows the mean search time of VOI-driven
solution count estimate deployment $T_{VSC}$ normalized by the search
time of exhaustive deployment $T_{SC}$ ($\gamma=0$), for the minimum
conflicts heuristic $T_{MC}$, and for the pAC heuristic $T_{PAC}$.
The shortest search time on average is achieved by VSC for $\gamma \in
[10^{-4},3\cdot10^{-3}]$ (shaded in the figure) and is much shorter
than for SC ($\mean \left(\frac {T_{VSC}(10^{-3})}
{T_{SC}}\right)\approx 0.45$); the improvement is actually close to
getting all the information provided by the heuristic without paying
the overhead at all. For all but one of the 14 benchmarks the search
time for VSC with $\gamma=3\cdot10^{-3}$ is shorter than for MC. For
most values of $\gamma$, VSC gives better results than MC ($\frac
{T_{VSC}} {T_{MC}} < 1$). pAC always results in the longest search
time due to the computational overhead.

Figure~\ref{fig:benchmarks}.b shows the mean number of backtracks of
VOI-driven deployment $N_{VSC}$ normalized by the
number of backtracks of exhaustive deployment $N_{SC}$,
the minimum conflicts heuristic $N_{MC}$, and for
the pAC heuristic $\overline N_{pAC}$. VSC causes less backtracking
than MC for $\gamma\le 3\cdot10^{-3}$ ($\frac {N_{VSC}}  {N_{MC}} <
1$). pAC always causes less backtracking than other heuristics, but
has overwhelming computational overhead.

Figure~\ref{fig:benchmarks}.c shows $C_{VSC}$, the number of
estimated solution counts of VOI-driven deployment, normalized by the
number of estimated solution counts of exhaustive deployment
$C_{SC}$. When $\gamma=10^{-3}$ and the best search time is achieved,
the solution counts are estimated  only in a relatively small number
of search states: the average number of estimations is ten times smaller than in the
exhaustive case ($\mean\left(\frac {C_{VSC}(10^{-3})}
{C_{SC}}\right)\approx 0.099$, $\median\left(\frac {C_{VSC}(10^{-3})}
{C_{SC}}\right)\approx 0.048$). 

The results show that although the solution counting heuristic
may provide significant improvement in the search time, further improvement is
achieved when the solution count is estimated only in a small fraction of
occasions selected using rational metareasoning.

\begin{figure}[h] 
\centering
\includegraphics[scale=0.67]{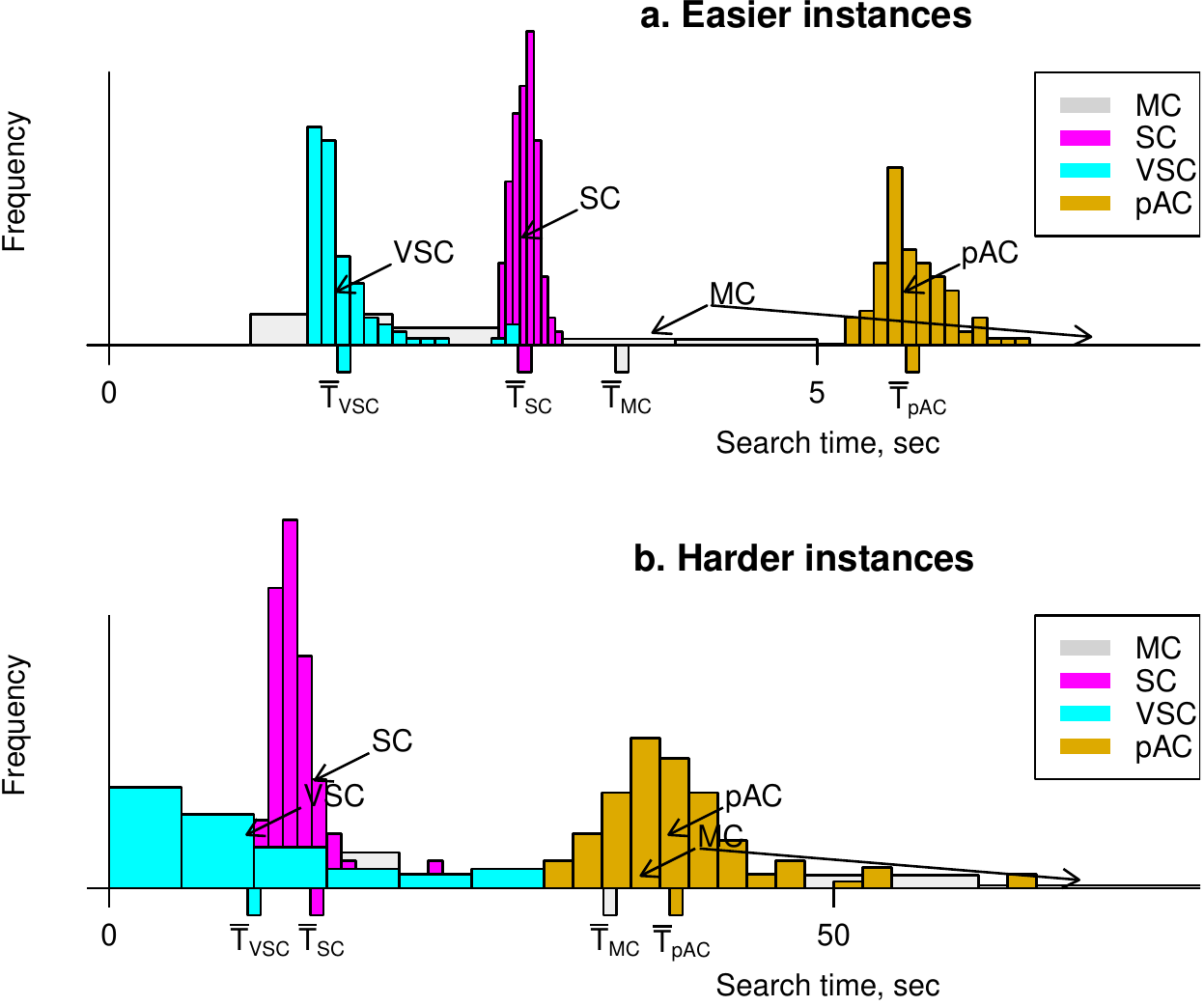}
\caption{Search time comparison on sets of random instances}
\label{fig:random-problems}
\end{figure}

\subsection{Random instances}

Based on the results on benchmarks, we chose $\gamma=10^{-3}$, and
applied it to two sets of 100 problem instances. 
Exhaustive deployment, rational deployment, the
minimum conflicts heuristic, and probabilistic arc consistency
were compared.

The first, easier, set was generated with 30 variables, 30 values per
domain, 280 constraints, and 220 nogoods per constraint.
Search time distributions are presented in
Figure~\ref{fig:random-problems}.a. The shortest mean search time is
achieved for rational deployment, with exhaustive deployment
next ($\frac {\overline T_{SC}} {\overline T_{VSC}}
\approx 1.75 $), followed by the minimum conflicts heuristic ($\frac
{\overline T_{MC}} {\overline T_{VSC}} \approx 2.16$) and 
probabilistic arc consistency ($\frac {\overline T_{pAC}} {\overline
  T_{VSC}} \approx 3.42$). Additionally,
while the search time distributions for solution counting are sharp
($\frac {\max T_{SC}} {\overline T_{SC}} \approx 1.08$, $\frac {\max
T_{VSC}} {\overline T_{VSC}} \approx 1.73$), the distribution for the
minimum conflicts heuristic has a long tail with a much longer worst
case time ($\frac {\max T_{VSC}} {\overline T_{VSC}} \approx 5.67$).

The second, harder, set was generated with 40 variables, 19 values,
410 constraints, 90 nogood pairs per constraint. 
Search time distributions are presented in Figure~\ref{fig:random-problems}.b.
As with the first set, the shortest mean search time is achieved for
rational deployment: $\frac {\overline T_{SC}} {\overline T_{VSC}}
\approx 1.43 $,  while the relative mean search time for the minimum
conflicts heuristic is much longer: $\frac {\overline T_{MC}}
{\overline T_{VSC}} \approx 3.45$. The probabilistic arc consistency heuristic
resulted again in the longest search time due to the overhead of computing
relative solution count estimates by loopy belief propagation: $\frac
{\max T_{VSC}} {\overline T_{VSC}} \approx 3.91$. 

Thus, the value of $\gamma$ chosen based on a small set of hard instances
gives good results on a set of instances with different parameters and
of varying hardness.

\subsection{Generalized Sudoku}

Randomly generated problem instances have played a key role in the
design and study of heuristics for CSP. However, one might argue that the
benefits of our scheme are specific to model RB. Indeed, real-world problem
instances often have much more structure than random instances
generated according to Model RB. Hence, we repeated the experiments
on randomly generated Generalized Sudoku
instances \cite{Ansotegui.sudoku}, since this domain is highly
structured, and thus a better source of realistic problems with a controlled
measure of hardness.


The search was run on two sets of 100 Generalized Sudoku instances,
with 4x3 tiles and 90 holes and with 7x4 tiles and 357 holes, with
holes punched using the doubly balanced method
\cite{Ansotegui.sudoku}. The search was repeated on each instance with
the exhaustive solution-counting, VOI-driven solution counting (with
the same value of $\gamma=10^{-3}$ as for the RB model problems),
minimum conflicts, and probabilistic arc
consistency value ordering heuristics. Results
are summarized in Table~\ref{tbl:sudoku} and show that relative
performance of the methods on Generalized Sudoku is similar to the
performance on Model RB.
\begin{table}[h]
\begin{center}
\small
\begin{tabular}{r|c|c|c|c}
               & $\overline {T_{SC}}$, sec & $\overline {\left(\frac
                   {T_{VSC}} {T_{SC}}\right)}$ & $\overline
               {\left(\frac {T_{MC}} {T_{SC}}\right)}$ & $\overline
               {\left(\frac {T_{pAC}} {T_{SC}} \right)}$ \\  \hline
4x3, 90 holes &  1.809 & 0.755 & 1.278 & 22.421 \\  \hline
7x4, 357 holes & 21.328 & 0.868 & 3.889 & 3.826
\end{tabular}
\end{center}
\caption{Generalized Sudoku}
\label{tbl:sudoku}
\end{table}

\subsection{Deployment patterns}

One might ask whether trivial methods for selective deployment would work.  
We examined deployment patterns of VOI-driven SC with
($\gamma=10^{-3}$) on several instances
of different hardness. For all instances, the solution counts were estimated
at varying rates during all stages of the search, and the
deployment patterns differ between the instances, so a simple
deployment scheme seems unlikely.  

VOI-driven deployment also compares favorably to
random deployment. Table~\ref{tbl:voirnd} shows
performance of VOI-driven deployment for $\gamma=10^{-3}$ and of
uniform random deployment, with total number of solution count estimations
equal to that of the VOI-driven deployment.
For both schemes, the values for which solution
counts were not estimated were ordered randomly, and the search
was repeated 20 times.  The mean search time for the random deployment is
$\approx1.6$ times longer than for the VOI-driven deployment, and has
$\approx100$ times greater standard deviation.
\begin{table}[h]
\begin{center}
\small
\begin{tabular}{r|c|c|c}
               & $\mathrm{mean}(T)$, sec & $\mathrm{median}(T)$, sec & $\mathrm{sd}(T)$, sec \\ \hline
VOI-driven     & 19.841                  & 19.815                    & 0.188 \\ \hline
random         & 31.421                  & 42.085                    & 20.038  
\end{tabular}
\end{center}
\caption{VOI-driven vs. random deployment}
\label{tbl:voirnd}
\end{table}

\section{Discussion and related work}
\label{sec:discussion}

The principles of bounded rationality appear in
\cite{Horvitz.reasoningabout}. \cite{Russell.right}
provided a formal description of rational metareasoning and case
studies of applications in several problem domains.
A typical use of rational metareasoning in search is in finding which node
to expand, or in a CSP context determining a variable or value assignment.
The approach taken in this paper adapts these methods to
whether to spend the time to compute a heuristic.

Runtime selection of heuristics has lately been of interest, e.g.
deploying heuristics for planning \cite{Karpas.maxornot}. The approach
taken is usually that of {\em learning} which heuristics to deploy based on
features of the search state. Although our approach can also benefit from learning,
since we have a parameter that needs to be
tuned, its value is mostly algorithm dependent,
rather than problem-instance dependent. This simplifies learning 
considerably, as opposed to having to learn a classifier from scratch.
Comparing metareasoning techniques to learning techniques (or possibly a combination
of both, e.g. by learning more precise distribution models) is an interesting issue for future research.

Although rational metareasoning is applicable 
to other types of heuristics, solution-count estimation heuristics are
natural candidates for the type of optimization suggested in this paper.
\cite{Dechter.cspheur} first suggested solution
count estimates as a value-ordering heuristic (using propagation on trees)
for constraint satisfaction problems, refined in \cite{Meisels.solcount} to multi-path propagation.

\cite{HorschHavens.pac} used a value-ordering heuristic that estimated
relative solution counts to solve constraint satisfaction problems and
demonstrated efficiency of their algorithm (called pAC, probabilistic
Arc Consistency). However, the computational overhead of the heuristic was large,
and the relative solution counts were
computed offline. \cite{Kask.solcount} introduced a CSP algorithm
with a solution counting heuristic based on the Iterative Join-Graph Propagation
(IJGP-SC), and empirically showed performance advances over MAC in most cases.
In several cases IJGP-SC was still slower than MAC due to the computational
overhead. 

Impact-based value ordering
\cite{Refalo.impact} is another heavy informative heuristic.  
One way to decrease its overhead, suggested in
\cite{Refalo.impact}, is to learn the impact of an assignment by averaging the impact of
earlier assignments of the same value to the same
variable. Rational deployment of this heuristic by estimating the probability 
of backtracking based on the impact may be possible, an issue for future research.
\cite{Gomes.solcount} propose a technique that adds random generalized XOR
constraints and counts solutions with high precision, but at present requires {\em solving} CSPs,
thus seems not to be immediately applicable as a search heuristic.

The work presented in this paper differs from the above related
schemes in that it does not attempt to introduce new heuristics
or solution-count estimates. Rather, an ``off the shelf''
heuristic is deployed selectively based on value of information,
thereby significantly reducing the heuristic's ``effective''
computational overhead, with an improvement in 
performance for problems of different size and hardness.

In summary, this paper suggests a model for adaptive deployment of value
ordering heuristics in algorithms for constraint satisfaction
problems. As a case study, the model was applied to a value ordering
heuristic based on solution count estimates, and a steady improvement in the
overall algorithm performance was achieved compared to {\em always} computing
the estimates, as well as to other simple deployment tactics.
The experiments showed that for many problem instances the optimum
performance is achieved when solution counts are estimated only in a
relatively small number of search states.

\section*{Acknowledgments}
The research is partially supported by the IMG4 Consortium under the MAGNET
program of the Israeli Ministry of Trade and Industry, by Israel
Science Foundation grant 305/09, by the Lynne and William Frankel
Center for Computer Sciences, and by the Paul Ivanier Center for
Robotics Research and Production Management.

\bibliographystyle{named}
\bibliography{refs}

\end{document}